\documentclass[conference]{IEEEtran}
\IEEEoverridecommandlockouts
\usepackage{cite}
\usepackage{amsmath,amssymb,amsfonts}
\usepackage{algorithmic}
\usepackage{graphicx}
\usepackage{textcomp}
\usepackage{xcolor}
\def\BibTeX{{\rm B\kern-.05em{\sc i\kern-.025em b}\kern-.08em
    T\kern-.1667em\lower.7ex\hbox{E}\kern-.125emX}}

\usepackage{fancyhdr,graphicx,amsmath,amssymb}
\usepackage[ruled,vlined]{algorithm2e}
\include{pythonlisting}
\usepackage[margin=1in]{geometry}
\usepackage[english]{babel}
\usepackage{subcaption}
\usepackage{comment}
\usepackage[skins]{tcolorbox}
\usepackage{csquotes}
\usepackage[normalem]{ulem}

\usepackage{algorithmic}

\newcommand{\startpara}[1]{{\vskip1pt\noindent{\bf #1.}}}
\newcommand{\sectref}[1]{Section~\ref{#1}}
\newcommand{\figref}[1]{Figure~\ref{#1}}

\newcommand{\agref}[1]{Algorithm~\ref{#1}}

\begin{document}

\title{Towards Personalized Explanation of \\ Robot Path Planning via User Feedback
\thanks{}
}

\author{\IEEEauthorblockN{Kayla Boggess, Shenghui Chen, and Lu Feng
\IEEEauthorblockA{Department of Computer Science,
University of Virginia\\
Charlottesville, USA\\
Emails: \{kjb5we, sc9by, lf9u\}@virginia.edu}}}

\maketitle

\begin{abstract}
Prior studies have found that explaining robot decisions and actions helps to increase system transparency, improve user understanding, and enable effective human-robot collaboration. In this paper, we present a system for generating personalized explanations of robot path planning via user feedback. We consider a robot navigating in an environment modeled as a Markov decision process (MDP), and develop an algorithm to automatically generate a personalized explanation of an optimal MDP policy, based on the user preference regarding four elements (i.e., objective, locality, specificity, and corpus). In addition, we design the system to interact with users via answering users' further questions about the generated explanations. Users have the option to update their preferences to view different explanations. The system is capable of detecting and resolving any preference conflict via user interaction. The results of an online user study show that the generated personalized explanations improve user satisfaction, while the majority of users liked the system's capabilities of question-answering and conflict detection/resolution.

\end{abstract}


\section{Introduction}\label{sec:intro}


Prior studies (e.g., \cite{anjomshoae2019explainable,chen2020towards,wang2016impact,hayes2017improving}) have found that providing explanations about robots' decisions and actions helps to increase system transparency, improve user understanding, and enable effective human-robot collaboration.
Different users have various preferences about what should be included in explanations, depending on users' goals, interests, and background.
However, a recent survey~\cite{anjomshoae2019explainable} finds that little research has been conducted for the generation of personalized explanations.
To fill in this gap, we develop a system for generating personalized explanations of robot path planning via user feedback.





We consider a robot navigating in an environment modeled as a Markov decision process (MDP), which is a popular modeling formalism for robot planning~\cite{Thrun05}.
Inspired by the line of works in the verbalization of robot experiences using natural language~\cite{rosenthal2016verbalization,perera2016dynamic}, we formalize the notion of \emph{user preference} as a tuple with four elements: the
\emph{objective} of robotic planning,  
\emph{locality} (i.e., which segment of the path should be explained),
\emph{specificity} (i.e., what level of details to include in the explanation),
and \emph{corpus} (i.e., what vocabulary to use in the explanation).
We develop an algorithm to generate a personalized explanation of an optimal MDP policy based on a given user preference tuple.
The algorithm outputs an ordered list of sentences instantiated using a structured language template.

Insights from social sciences~\cite{miller2019explanation} show that humans prefer \emph{interactive} explanations (e.g., via user-system dialogues) and \emph{contrastive} explanations (e.g, ``Why A rather than B'').
Therefore, we design the system to interact with users via question-answering. 
After being presented with an explanation, the user can ask further (structured) questions to clarify the robotic plan.
For example, ``Why does the robot move east rather than taking a different action in the landmark?''
The system answers the user's question with a contrastive explanation: 
``The robot moves east in the landmark because it is part of the optimal robotic plan to achieve the mission objective, while taking a different action in the landmark cannot guarantee the mission objective.''

During the interaction with the system, users have the option to update their preferences (e.g., for viewing different explanations of the same robotic plan, or for re-planning with different objectives). 
The system checks if there is any conflict between the updated and the previous user preferences, and guides users step by step to resolve any preference conflict. 
Then, the system generates new explanations and/or plans based on the updated preferences. 
Finally, the system terminates once users stop choosing to update their preferences and the robotic planner is able to finalize the chosen route.

We adopt a commonly used subjective evaluation method to evaluate the proposed system via an online user study with 88 participants using Amazon Mechanical Turk. 
We asked users about their understanding and satisfaction with different types of explanations. 
As a baseline for comparison with personalized explanations, we also presented users with basic explanations (i.e., a statement of the robot's choice of actions without further reasoning) that are not customized based on user preferences.
In addition, we asked users if they liked the system's capabilities of question-answering and conflict resolution. 
The study results show that personalized explanations generated using our system can lead to better user satisfaction than basic explanations; and users reported a similar level of understandings about both types of explanations. 
82.95\% and 93.18\% of users indicated that they liked the system's question-answering and conflict-resolving capabilities, respectively.

\startpara{Contributions}
We summarize the major contributions of this work as follows.
\begin{itemize}
    \item We developed a system for generating personalized explanations of robot path planning based on user preferences.
    \item We equipped the system with capabilities to interact with users for feedback, answer users' questions about the generated explanations, and detect and resolve user preference conflicts.
    \item We designed and conducted an online user study to evaluate the proposed system, which showed encouraging results.
\end{itemize}

\startpara{Paper Organization}
The rest of the paper is organized as follows.
We survey the related work in \sectref{sec:related},
present the system for generating personalized explanations of robotic planning in \sectref{sec:approach},
describe the online user study design in \sectref{sec:study},
discuss the study results in \sectref{sec:results},
and draw conclusions in \sectref{sec:conclusion}.

\section{Related Work}\label{sec:related}

Explainable AI (XAI)~\cite{gunning2019xai} has been drawing increasing interest in recent years. 
The goal of an XAI system is to provide explanations about its decisions and actions for better transparency. 
The notions of explainability are domain-dependent. 
Many XAI systems for data-driven applications relate explainability with interpretability, and build simplified models that approximate complex (black-box) machine learning mechanisms (e.g., deep neural networks)~\cite{adadi2018peeking,arrieta2020explainable}.
By contrast, XAI systems for robots aim to explain robots' behavior and inform about robots' intents to human users. 
Existing works have presented explanations for robots in different formats:
text-based language (e.g., \cite{chen2020towards,wang2016impact,hayes2017improving}), graphs and images (e.g., \cite{chen2018situation}), and expressive lights (e.g., \cite{baraka2016expressive}).
We adopt the most commonly used text-based explanations. 

Specifically, we use structured language templates to instantiate explanations, which has been explored in several prior works.
A concept of verbalization was introduced in~\cite{rosenthal2016verbalization}, which enables robots to describe their experiences (e.g., navigation, perception) through structured language.
An autonomous policy explanation approach was presented in~\cite{hayes2017improving}, which synthesizes descriptions of robots' optimal policies and responds to users' queries instantiated using language templates.
In addition, an approach was proposed in~\cite{feng2018counterexamples} to describe counterexamples (e.g., illustrating why mission requirements are violated) of robotic planning using structured sentences.  
Our recent work~\cite{chen2020towards} investigated the automated generation of contrastive explanations represented as structured language. 

Nevertheless, as identified in a recent survey~\cite{anjomshoae2019explainable} on explainable robots, little research has been conducted for personalized explanations.
A user-aware approach was presented in~\cite{kaptein2017personalised} to customize explanations of robot actions based on user age (children or adults).
Context-aware explanations (i.e., selecting the best explanation to provide based on the context) have been explored for human-robot teaming~\cite{gong2018behavior} and robot navigation~\cite{hastie2018miriam}.
None of these works consider personalized explanations based on user preferences.

Personalized explanations based on user preferences have been investigated in other application domains, such as recommendation systems~\cite{quijano2017make}.
In addition, social media and recommendation systems have looked at the possibility of conflicts regarding user preferences. Social media preference conflicts occur when users disagree over preferences such as a post's share settings and are easily resolved by following one user's preference based on the situation~\cite{SocialMediaConflicts,SocialMediaConflicts2,SocialMediaConflicts3}. Recommendation system conflicts arise when no product or action matches a user's preferences and resolution occurs by explaining why a choice is shown or recommending a new preference to set~\cite{RecommendationSystemConflicts,Pu2008UserInvolvedPE}. However, there has been little work to detect conflicts within the user preferences themselves, especially for robotic planning explanations, as these types of personalized explanation conflicts are often more complex, involving multiple preferences and possibly resolving in an unsatisfying way.

\section{Approach} \label{sec:approach}

\begin{figure}[t]
     \centering
     \includegraphics[width=\columnwidth]{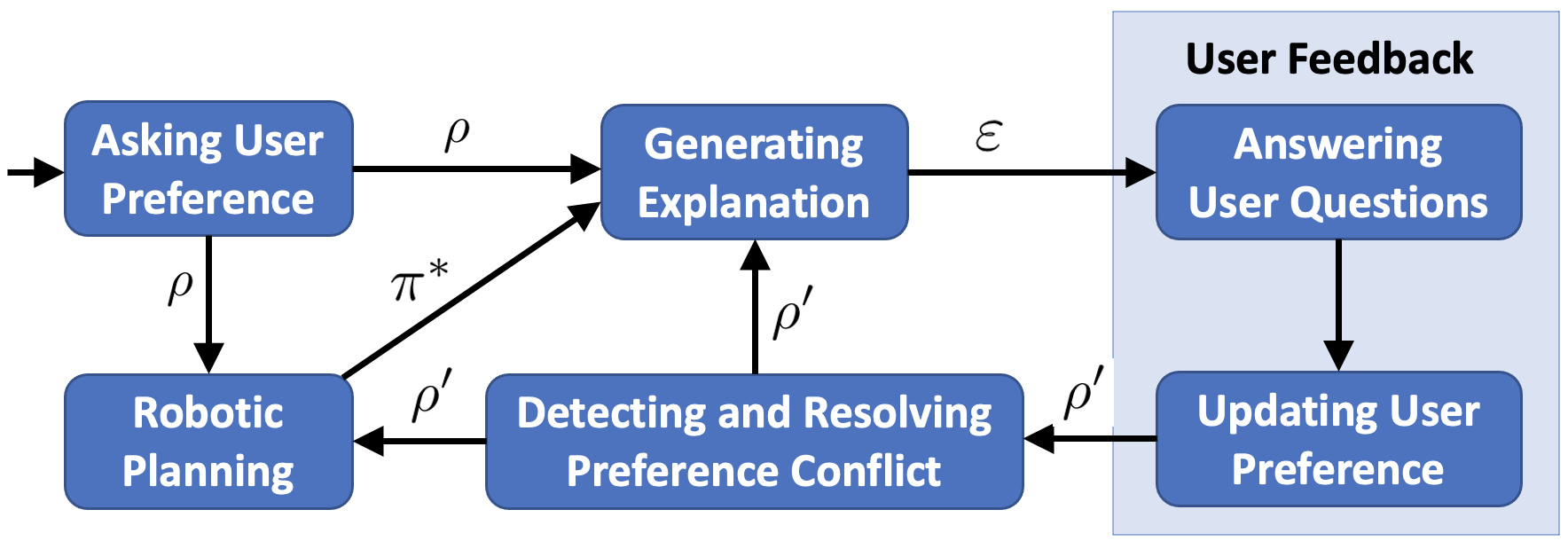}
     \caption{An overview of the proposed system. }
     \label{fig:flowchart}
 \end{figure}

We develop a system for generating personalized explanations of robot path planning via user feedback, as illustrated in \figref{fig:flowchart}.
After obtaining the initial user preference $\rho$ (\sectref{sec:preference}), the system computes an optimal robotic plan $\pi^*$ based on the objective specified in the user preference (\sectref{sec:planning}), and generates an explanation $\varepsilon$ of the robotic plan based on the user preference (\sectref{sec:explanation}).  
Next, the system interacts with the user for feedback, including answering the user's questions about the robotic plan and, if the user desires, updating the user preference (\sectref{sec:feedback}). 
The system detects and resolves any preference conflict (\sectref{sec:conflict}). 
Based on the updated preference $\rho'$, the system computes a new robotic plan and generates an explanation.
This process iterates until the user stops choosing to update the user preference. 
At this point, the robot executes the finalized plan.

\subsection{User Preference}\label{sec:preference}

We use a tuple $\rho = (ob,lo,sp,co)$ to represent the user preference, which includes the following four elements:
\begin{itemize}
    \item \textbf{Objective} (denoted $ob$) of the robotic planning (e.g., finding the shortest/safest route). 
    \item \textbf{Locality} (denoted $lo$) describes the segment(s) of the robotic plan that the user is interested in. The user may want to know the robotic plan in the global environment (e.g., the entire map), or only part of it (e.g., a region of the map).
    \item \textbf{Specificity} (denoted $sp$) indicates the level of detail to include in the explanation: a summary of the robotic behavior in a pre-defined set of critical states (e.g., states representing landmarks), or a detailed narrative about every state.
    \item \textbf{Corpus} (denoted $co$) determines the vocabulary of the explanation to describe the robot's concrete state-based world representation (e.g., \emph{move west} in \emph{grid 12}) or high-level representation (e.g., \emph{move along the corridor} in \emph{the landmark}).
\end{itemize}
Note that the above definitions of locality, specificity, and corpus are inspired and adapted from~\cite{rosenthal2016verbalization}. 
There are many different ways to elicit user preferences of these elements, for example, 
asking users to select from a predetermined list of keywords and examples.
Users can update their preferences at any time before the robot begins to execute the chosen plan.

\subsection{Robotic Planning}\label{sec:planning}

We consider robot path planning based on MDPs, which can be denoted as a tuple $(S, s_0, A, \delta, R)$, 
where $S$ is a finite set of states, 
$s_0\in S$ is an initial state, 
$A$ is a set of actions,
$\delta: S\times A\times S \to [0,1]$ is a probabilistic transition function, 
and $R: S\times A \times S \to \mathbb{R}$ is a real-valued reward function.
The goal of planning is to compute an optimal policy $\pi^*: S \to A$ for a given objective specified in the user preference $\rho$.
There are various MDP planning methods.
For example, reinforcement learning~\cite{sutton2018reinforcement} can learn an optimal policy that maximizes the cumulative reward $\mathbb{E} [\sum_{t=0}^\infty \gamma^t R(s_t, a_t, s_{t+1})]$, where $\gamma$ is a discount factor. 
The proposed system is agnostic to the choice of planning methods and can be used to explain any MDP policy in general.

\begin{figure}[t]
     \centering
     \includegraphics[width=0.6\columnwidth]{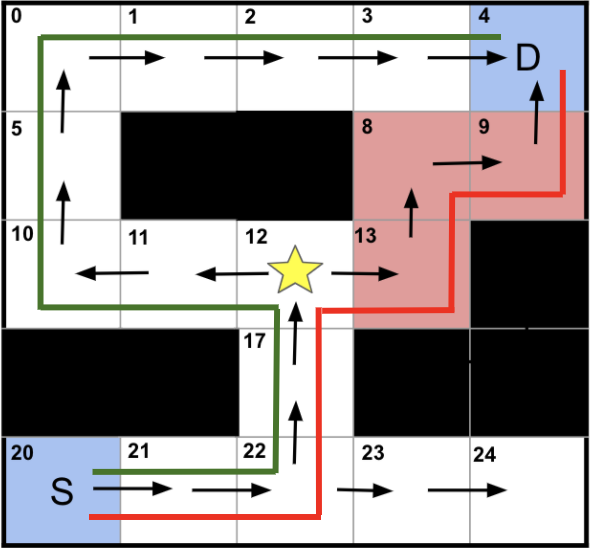}
     \caption{A map for robotic planning. Blue: start (S) and destination (D). Black: obstacles. Star: landmark. Red grids: crowded passage. White grids: corridor.}
     \vspace{-10pt}
     \label{fig:map}
\end{figure}

We use a grid-world example to illustrate the proposed approach. 
Consider the route planning for a robot navigating in the $5\times5$ grid map shown in \figref{fig:map}.
We build an MDP model with 25 states. Each state corresponds to a grid in the map.
The starting state $s_0$ is grid 20 (labeled with S). 
The action space is given by the robot's navigation directions (shown as arrows in Figure \ref{fig:map}). 
We assume that the robot can move to the next grid in its intended direction with a probability of 0.8. Otherwise, the robot will stay put or move in other directions with a combined probability of 0.2.
We design two reward functions: a distance reward $R_1$ that counts the negation\footnote{Note that negations are needed here because reinforcement learning maximizes the cumulative reward, and we want to minimize the total distance (\emph{resp.} safety risk) for the shortest (\emph{resp.} safest) route.} of the number of navigated grids, and a safety reward $R_2$ that counts the negation of the number of navigated red grids (representing crowded passage where the robot is more likely to collide with pedestrians). 
Suppose that the user gives an objective of finding the shortest route for the robot to navigate from the starting location (grid 20) to the destination (grid 4).
The resulting optimal route is the red route in the map.
Suppose that the user's objective is to find the safest route for the robot. The resulting optimal route is the green route in the map, which does not pass through any crowded passage (no red grids) despite being longer in distance than the red route.

\begin{algorithm}[t]
\caption{Generating explanation sentences} \label{code}
\algsetup{linenosize=\tiny}
\small
\begin{algorithmic} [1]
\REQUIRE A user preference $\rho = (ob,lo,sp,co)$, an optimal robotic plan $\pi^*$, 
    an MDP $(S, s_0, A, \delta, R)$, a structured language template $\tau$
\ENSURE An ordered list of sentences $\varepsilon$
\STATE $\varepsilon \leftarrow \{\}$
\STATE $S' \leftarrow \mathsf{findStates}(S, lo, sp)$
\STATE $V \leftarrow \mathsf{findCorpus}(co)$
\STATE $Q \leftarrow \{s_0\}$
\WHILE{$Q$ is non-empty}
    \STATE remove a state $s$ from the head of $Q$
    \STATE find the action $a \leftarrow \pi^*(s)$
    \FORALL{state $s'$ with $\delta(s,a,s') > 0$}
        \STATE insert $s'$ to the tail of Q
    \ENDFOR
    \IF{$s \in S'$}
        \STATE instantiate a sentence $e \leftarrow \tau(V,s,a)$
        \STATE add $e$ to the tail of $\varepsilon$
    \ENDIF
\ENDWHILE
\RETURN $\varepsilon$
\end{algorithmic}
\end{algorithm}

\subsection{Generating Explanation}\label{sec:explanation}

\agref{code} illustrates the procedure of generating a personalized explanation $\varepsilon$ (i.e., an ordered list of sentences describing the agent's potential actions) of an optimal robotic plan $\pi^*$ based on the user preference $\rho$. 
The algorithm uses a $\mathsf{findStates}(S, lo, sp)$ function (line 2) to identify the set of MDP states $S'$ that need to be explained, based on the user preference of locality $lo$, i.e., which segment(s) of the robotic plan that the user is interested in, and specificity $sp$, i.e., the level of details that the user wants to know (every state or predefined critical state).
The algorithm also uses a $\mathsf{findCorpus}(co)$ function (line 3) to determine the vocabulary $V$ for the explanation, given the user preference of corpus $co$ (concrete or high-level world representation).
The algorithm uses a queue $Q$ for the topological sorting of MDP states (line 4-10).
For each state $s \in S'$, the algorithm instantiates a sentence $e$ using a structured language template  $\tau(V,s,a)$. This template is chosen previously by the developer to match the needs of the given agent and environment. An example for our grid world is as follows:
\begin{center}
The robot $\langle$action $a\rangle$ in $\langle$state $s\rangle$.
\end{center}
The template is instantiated with words in the vocabulary $V$ to describe the state $s$ and the action $a=\pi^*(s)$ given by the robotic plan. 
The instantiated sentence $e$ is then added to an ordered list $\varepsilon$.
The algorithm terminates when $Q$ is empty, and returns the ordered list of sentences $\varepsilon$ as an explanation of the robotic plan. 

Following the robot navigation example described in \sectref{sec:planning}, 
suppose that the user gives a preference $\rho=(ob,lo,sp,co)$ with the objective $ob$ of finding the shortest route, the locality $lo$ of considering the route segment between the landmark and the destination, the specificity $sp$ of describing every state, and the corpus $co$ of using the vocabulary about the robot's concrete world representation. 
Based on this user preference, \agref{code} generates the following ordered list of sentences as a personalized explanation of the optimal robotic plan (red route in \figref{fig:map}):
\vspace{10pt}
\begin{enumerate}
    \item The robot moves east in grid 12.
    \item The robot moves north in grid 13.
    \item The robot moves east in grid 8. 
    \item The robot moves north in grid 9. 
    \item The robot stops in grid 4.
\end{enumerate}

Suppose that the user gives a different preference $\rho'=(ob',lo',sp',co')$ with the objective $ob'$ of finding the safest route, the locality $lo'$ of considering the entire route, the specificity $sp'$ of including only critical states (start, landmark, destination), and the corpus $co'$ of using the vocabulary about the high-level world representation. 
Based on the new user preference, \agref{code} generates the following ordered list of sentences as a personalized explanation of the optimal robotic plan (green route in \figref{fig:map} where the corridor is represented as grids in white color):
\begin{enumerate}
    \item The robot moves along the corridor in the start. 
    \item The robot moves along the corridor in the landmark.
    \item The robot stops in the destination.
\end{enumerate}

\subsection{User Feedback}\label{sec:feedback}

According to insights from the social sciences~\cite{miller2019explanation}, a good explanation shall be interactive (e.g., via a dialogue between the user and the system) and contrastive (e.g., explaining ``Why A rather than B'').
Therefore, we design the proposed system to be able to supplement the previously given explanation and interact with the user for contrastive feedback (further clarification of the robot's reasoning or why the robot performs an action) about the robotic plan. This way, the user is provided only with the information that they request, and the cognitive burden of unneeded information is reduced.

Specifically, after showing the user an explanation generated by \agref{code}, the system allows the user to ask questions about the explained robotic plan using the following structured language template: 
\begin{displayquote}
Why does the robot $\langle$action$\rangle$ rather than take a different action in $\langle$state$\rangle$?
\end{displayquote}
The user can instantiate the question template with words describing the robot's action and state taken from the displayed explanation sentences. 
The system then answers the question using a contrastive explanation with the following structured language template: 
\begin{displayquote}
The robot $\langle$action$\rangle$ rather than taking a different action in $\langle$state$\rangle$, because it is part of the optimal robotic plan to achieve the $\langle$objective$\rangle$, while taking a different action in $\langle$state$\rangle$ cannot guarantee the $\langle$objective$\rangle$. 
\end{displayquote}

For example, the user asks ``Why does the robot move east rather than taking a different action in grid 12''? 
The system answers with a contrastive explanation:   
``The robot moves east in grid 12, because it is part of the optimal robotic plan to achieve the shortest route, while taking a different action in grid 12 cannot guarantee the shortest route.''

The interaction between the user and the system continues until the user has no further questions. At this point, the system asks if the user would like to update the current preferences. 
The system then checks if there is any preference conflict as follows.

\subsection{Preference Conflict Detection and Resolution}\label{sec:conflict}

The system detects and resolves any \emph{conflict} (detected difference from previous preferences) between the updated user preference $\rho'=(ob',lo',sp',co')$ and the previous user preference $\rho=(ob,lo,sp,co)$ using various rules as users can update their preferences any time before route finalization. The conflicts are as follows:
\begin{itemize}
    \item \textbf{Soft conflict} if the objective remains the same ($ob' = ob$) but any other elements of the preference tuple change ($lo' \neq lo \vee sp' \neq sp \vee co' \neq co$). The system confirms if the user just wants to view a different explanation of the same robotic plan, assuming that the same objective of robotic planning results in the same plan. If the user replies ``yes'', then the system uses \agref{code} to generate a new personalized explanation based on $\rho'$, and follows with the user feedback interaction as described in \sectref{sec:feedback}. If the user replies ``no'', then the system assumes that the user mistakenly entered their preferences incorrectly, and the system asks the user to update $\rho'$ to reflect the intended changes.  It is important to note any soft conflict causing elements cannot conflict with one another as they are only cosmetic changes to the explanation that affect how the information is presented to the user.
    \item \textbf{Hard conflict} if the preference of the objective changes ($ob' \neq ob$). The system confirms if the user indeed wants to update the planning objective. If the user replies ``yes'', then the system computes a new optimal robotic plan based on the updated objective $ob'$ and generates a new personalized explanation about the new plan based on $\rho'$. If the user replies ``no'', the system assumes a mistake, and the system asks the user to revise $\rho'$.
    \item \textbf{No conflict} if $\rho'=\rho$. The system interacts with the user to confirm that the user has finished updating. If the user replies ``no'', then the user needs to provide a new preference different from $\rho'$. If the user replies ``yes'', the system terminates. The user can update their preferences as many times as they wish before the route is finalized by the planner.
\end{itemize}

\section{User Study Design} \label{sec:study}

We designed and conducted an online user study to evaluate the proposed approach.  
We describe the study design here and analyze the results in \sectref{sec:results}.

\startpara{Experiment Design} 
We recruited 100 individuals with a categorical age distribution of 2 (18-24); 49 (25-34); 26 (35-49); 9 (50-64); and 2 (65+) using Amazon Mechanical Turk (AMT). To ensure data quality, we only accepted users on AMT that had a 90\% approval rate and had performed at least 50 tasks previously. Additionally, we injected attention check questions periodically during the study.
After filtering out low-quality answers, we end up using data from 88 participants for the analysis.

We instructed each user to consider a robot path planning problem in a grid map environment similar to Figure 2. Users were briefed about how the robot can move (north, south, east, west, stop) and the types of states that it may encounter (start, destination, obstacles, landmarks, corridors, etc.). We started by eliciting user preferences for all four elements (objective, locality, specificity, and corpus) by having the user choose from a drop-down list with explanation examples for each element. We then derived the user’s preference tuple and applied it to synthesize a personalized explanation for nine different grid map examples (3 states over 3 maps). Once the explanation was synthesized for a map and the questions about it submitted, the user was able to update their preferences for the next map (as many times as they liked from a drop-down) if they were not happy with the state of the provided information and continue on with these new preferences. We randomized the order of the presented explanations and the maps to counterbalance the ordering confound effect. 

\startpara{Manipulated Factors}
We manipulated one factor: explanation type. For each map, a personalized explanation (e.g., ``We move up because it leads to the shortest route.'') with the user’s current preferences was presented alongside a basic explanation of a single statement of the robot’s choice of action without further reasoning (e.g., ``We move north towards the destination.'') 

\startpara{Dependent Measures}
For each explanation type, we asked users to rate on a 5-point Likert item their understanding (5 for fully understand and 1 for no understanding) and satisfaction (5 for fully satisfied and 1 for not satisfied). We then asked users to select the explanation that was more favorable to them for each map and also overall at the end of the survey. Finally, we kept track of the number of times that the user updated their preferences to modify the preference-based explanation for each of the nine maps.

\startpara{Hypotheses}
\begin{itemize}
    \item \textbf{H1:} Personalized explanations are more understandable than basic explanations.
    \item \textbf{H2:} Personalized explanations are more satisfying than basic explanations.
    \item \textbf{H3:} Personalized explanations are more favorable than basic explanations.
\end{itemize}

\section{Results} \label{sec:results}

\begin{table*}[]
\resizebox{\textwidth}{!}{%
\begin{tabular}{cccccccclcc}
\hline
\multicolumn{2}{c}{Objective} &  & \multicolumn{2}{c}{Locality} &  & \multicolumn{2}{c}{Specificity} &  & \multicolumn{2}{c}{Corpus} \\ \cline{1-2} \cline{4-5} \cline{7-8} \cline{10-11} 
Shortest Path & 45 &  & Global & 67 &  & All Information & 69 &  & Lefts and Rights & 77 \\
Safest Path & 27 &  & Only Highways & 15 &  & Important Information & 19 &  & Landmarks & 11 \\
Shortest and Safest Path & 16 &  & Only Alleyways & 2 &  &  &  &  &  &  \\
 &  &  & Single Position & 4 &  &  &  &  &  &  \\ \hline
\end{tabular}%
}
\caption{User selection of preference elements options for personalized explanations }
\label{tab:preferences}
\end{table*}

\begin{figure}[t]
     \centering
     \includegraphics[width=0.95\columnwidth]{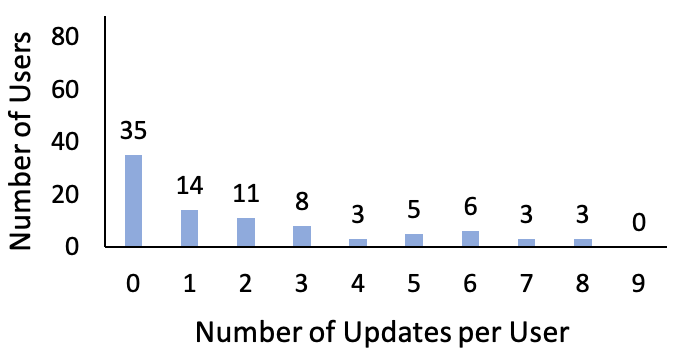}
     \caption{Number of preference updates per user vs. number of users in each category.}
     \vspace{-10pt}
     \label{fig:updatefiguredraft}
 \end{figure}
 

 
 \begin{figure}[t]
     \centering
     \includegraphics[width=1.0\columnwidth]{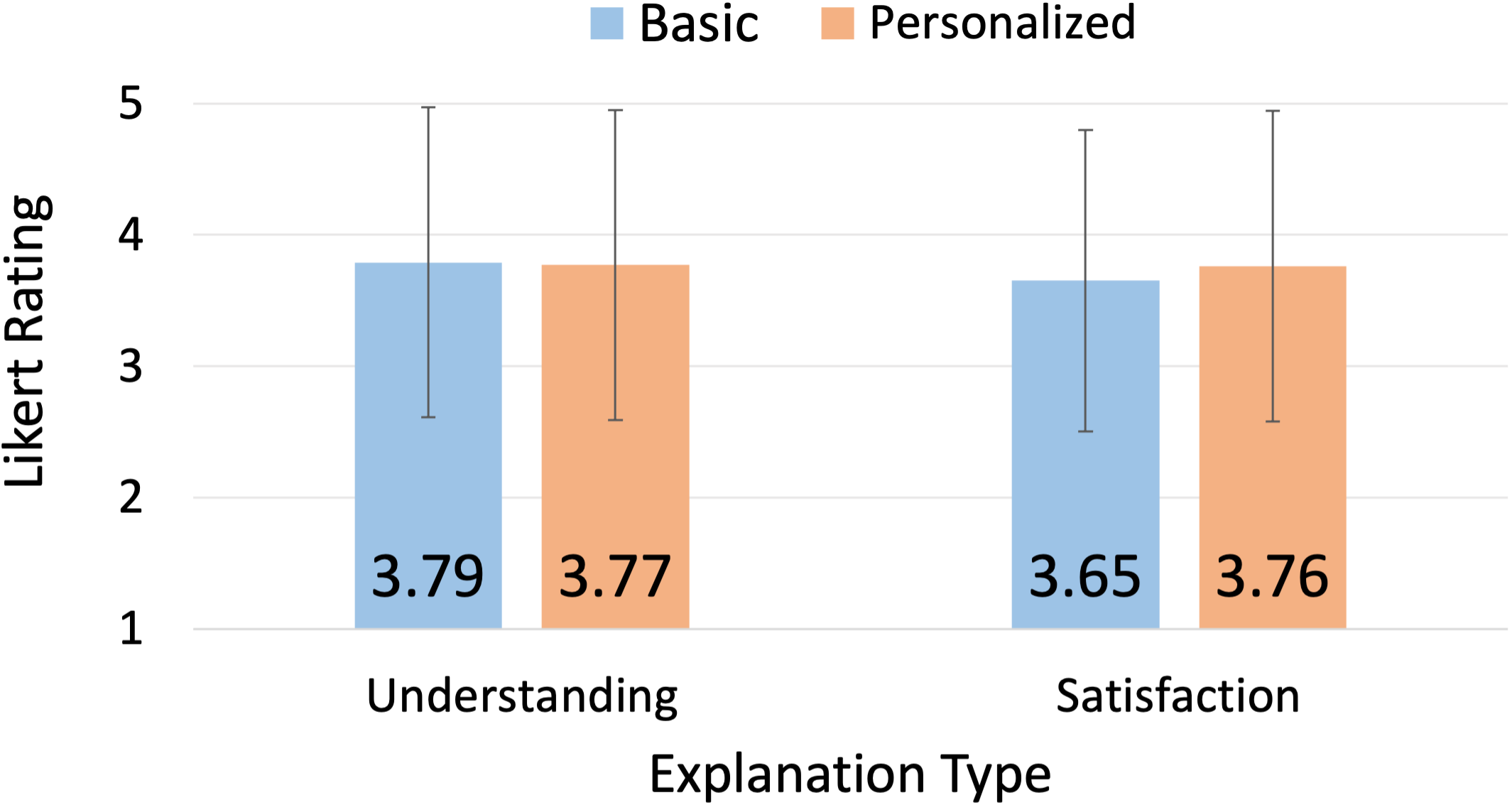}
     \caption{User understanding and satisfaction (mean and standard deviation) about basic and personalized explanations.}
     \vspace{-10pt}
     \label{fig:combinedchart}
 \end{figure}

\startpara{Preference elicitation and updates} 
As shown in Table \ref{tab:preferences}, we observe that users in the online study selected preferences across all available options objective, locality, specificity, and corpus. This observation confirms our assumption that user preferences are diverse, and thus important to consider in explanations to meet user expectations.
In addition, we observed a total of 178 preference updates among 88 users. Figure \ref{fig:updatefiguredraft} shows a breakdown of the numbers: 35 users chose not to update their preferences, while 53 users (60.23\% of the study group) updated their preferences at least once during the study. This indicates that most users will update their preferences to match a given route or map if the option is given and shows the necessity of allowing for the update of preferences to occur. 

\startpara{Statistical analysis for hypotheses} 
To evaluate H1 and H2, we first begin by testing the normality of our responses using a Shapiro-Wilk test (Understanding:$\alpha$= 0.05, W = 0.855 , p$\leq0.000$, Skewed. Satisfaction:$\alpha$= 0.05, W = 0.870 , p$\leq0.000$, Skewed.). Finding that neither set of data is normally distributed, we employ a Friedman test (a repeated measure version of the more common Kruskal-Wallis test) \cite{daniel1990applied} to prove the statistical significance of the means of all responses between personalized and basic explanations.  For each Friedman test, we use an alpha value of 0.05 to ensure a confidence level of 95\% and assume a null hypothesis that the mean of responses for both explanation types will be statistically similar. 
As shown in Figure \ref{fig:combinedchart}, we find that there is no significant statistical difference in user understanding between explanation types (Friedman: $\alpha$= 0.05, Q = 0.456 , p$\leq$0.500, No Significant Difference.). This is most likely due to both explanations having a goal of imparting all necessary information to the user and the simplicity of maps used in the study. \textbf{So, we reject H1.}

Figure \ref{fig:combinedchart} shows that we find that users are more satisfied with preference-based explanations than their basic counterparts (Friedman: $\alpha$= 0.05, Q = 6.365 , p$\leq$0.012, Significant Difference.). \textbf{So, we accept H2.}

To evaluate H3, we utilize a Pearson's chi-squared test \cite{seltman2012experimental} to prove the statistical significance in the frequency of explanation selection. We assume an expected selection of 50/50 to represent a random selection of explanations by the users and assume a null hypothesis that the user selection of explanations will be random/fit our expected values. We once again use an alpha value of 0.05. When asked to choose which type of explanations they preferred, users choose the preference-based explanation 56.95\% of the time ($\chi^2$ :$\alpha$= 0.05, $\chi^2$ = 15.278, $\chi^2_c$ = 3.84, p$\leq$0.000093, Reject $H_0$.). Overall, 92.05\% indicate they prefer preference-based explanations to basic explanations ($\chi^2$ :$\alpha$= 0.05, $\chi^2$ = 62.23, $\chi^2_c$ = 3.84, p$\leq$0.00001, Reject $H_0$.). This higher level of satisfaction and favor may be caused by the perceived increase in user agency (i.e., the ability of users to make choices and act independently \cite{emirbayer_mische_1998}), since the users were able to choose the aspects of the explanations themselves. Additionally, by choosing the factors themselves, users were able to better guarantee that they would “like” the explanations more as they were able to request more/different information if they needed it, but did not have to deal with the additional burden of information if it was not required. \textbf{So, we accept H3.}

\startpara{Question-answering and conflict detection} 
We also gave each user the chance to use the interactive (answering further questions) and conflict detection portion of our system. First, each user was allowed to ask further questions (from a predefined list) about the personalized explanations that they received. 
We asked each user if overall they liked the ability to ask further questions/receive further reasoning about the robot’s actions and 82.95\% of users stated that they did like this ability. 
Then each user was given 3 example preference conflicts, one soft (locality updated from only highways to only alleyways), one hard (shortest path updated to safest path), and one none (no updates), to resolve using the process described in our approach. We asked each user if they liked the preference conflict resolver and 93.18\% of users stated that they did. Further testing will need to be done to support the use of these abilities over other possible explanation options.

\section{Conclusion} \label{sec:conclusion}
In this paper, we present a system for generating personalized explanations of robotic planning based on user preferences. 
The system also has the capabilities to interact with users for feedback, answer users' questions about the generated explanations, allow users to update their preferences, and detect and resolve any user preference conflict.
The results of our user study show that customizing explanations based on user preferences increases user satisfaction, but does not improve user understanding of explanations.
Furthermore, the majority of users in our study indicated that they liked the system's question-answering capability and conflict detection/resolution capability, respectively.
In the future, we plan to extend this approach to generate personalized explanations for multi-user, multi-agent applications.

\bibliographystyle{./bibliography/IEEEtran}
\bibliography{reference}

\begin{thebibliography}{10}
\providecommand{\url}[1]{#1}
\csname url@samestyle\endcsname
\providecommand{\newblock}{\relax}
\providecommand{\bibinfo}[2]{#2}
\providecommand{\BIBentrySTDinterwordspacing}{\spaceskip=0pt\relax}
\providecommand{\BIBentryALTinterwordstretchfactor}{4}
\providecommand{\BIBentryALTinterwordspacing}{\spaceskip=\fontdimen2\font plus
\BIBentryALTinterwordstretchfactor\fontdimen3\font minus
  \fontdimen4\font\relax}
\providecommand{\BIBforeignlanguage}[2]{{%
\expandafter\ifx\csname l@#1\endcsname\relax
\typeout{** WARNING: IEEEtran.bst: No hyphenation pattern has been}%
\typeout{** loaded for the language `#1'. Using the pattern for}%
\typeout{** the default language instead.}%
\else
\language=\csname l@#1\endcsname
\fi
#2}}
\providecommand{\BIBdecl}{\relax}
\BIBdecl

\bibitem{anjomshoae2019explainable}
S.~Anjomshoae, A.~Najjar, D.~Calvaresi, and K.~Fr{\"a}mling, ``Explainable
  agents and robots: Results from a systematic literature review,'' in
  \emph{18th International Conference on Autonomous Agents and Multiagent
  Systems (AAMAS 2019), Montreal, Canada, May 13--17, 2019}.\hskip 1em plus
  0.5em minus 0.4em\relax International Foundation for Autonomous Agents and
  Multiagent Systems, 2019, pp. 1078--1088.

\bibitem{chen2020towards}
S.~Chen, K.~Boggess, and L.~Feng, ``Towards transparent robotic planning via
  contrastive explanations,'' in \emph{Proceedings of the International
  Conference on Intelligent Robots and Systems (IROS)}, 2020.

\bibitem{wang2016impact}
N.~Wang, D.~V. Pynadath, and S.~G. Hill, ``The impact of pomdp-generated
  explanations on trust and performance in human-robot teams,'' in
  \emph{Proceedings of the 2016 international conference on autonomous agents
  \& multiagent systems}, 2016, pp. 997--1005.

\bibitem{hayes2017improving}
B.~Hayes and J.~A. Shah, ``Improving robot controller transparency through
  autonomous policy explanation,'' in \emph{2017 12th ACM/IEEE International
  Conference on Human-Robot Interaction (HRI}.\hskip 1em plus 0.5em minus
  0.4em\relax IEEE, 2017, pp. 303--312.

\bibitem{Thrun05}
S.~Thrun, W.~Burgard, and D.~Fox, \emph{Probabilistic Robotics}.\hskip 1em plus
  0.5em minus 0.4em\relax The MIT Press, 2005.

\bibitem{rosenthal2016verbalization}
S.~Rosenthal, S.~P. Selvaraj, and M.~M. Veloso, ``Verbalization: Narration of
  autonomous robot experience.'' in \emph{IJCAI}, vol.~16, 2016, pp. 862--868.

\bibitem{perera2016dynamic}
V.~Perera, S.~P. Selveraj, S.~Rosenthal, and M.~Veloso, ``Dynamic generation
  and refinement of robot verbalization,'' in \emph{2016 25th IEEE
  International Symposium on Robot and Human Interactive Communication
  (RO-MAN)}.\hskip 1em plus 0.5em minus 0.4em\relax IEEE, 2016, pp. 212--218.

\bibitem{miller2019explanation}
T.~Miller, ``Explanation in artificial intelligence: Insights from the social
  sciences,'' \emph{Artificial Intelligence}, vol. 267, pp. 1--38, 2019.

\bibitem{gunning2019xai}
D.~Gunning, M.~Stefik, J.~Choi, T.~Miller, S.~Stumpf, and G.-Z. Yang,
  ``Xai—explainable artificial intelligence,'' \emph{Science Robotics},
  vol.~4, no.~37, 2019.

\bibitem{adadi2018peeking}
A.~Adadi and M.~Berrada, ``Peeking inside the black-box: A survey on
  explainable artificial intelligence (xai),'' \emph{IEEE Access}, vol.~6, pp.
  52\,138--52\,160, 2018.

\bibitem{arrieta2020explainable}
A.~B. Arrieta, N.~D{\'\i}az-Rodr{\'\i}guez, J.~Del~Ser, A.~Bennetot, S.~Tabik,
  A.~Barbado, S.~Garc{\'\i}a, S.~Gil-L{\'o}pez, D.~Molina, R.~Benjamins
  \emph{et~al.}, ``Explainable artificial intelligence (xai): Concepts,
  taxonomies, opportunities and challenges toward responsible ai,''
  \emph{Information Fusion}, vol.~58, pp. 82--115, 2020.

\bibitem{chen2018situation}
J.~Y. Chen, S.~G. Lakhmani, K.~Stowers, A.~R. Selkowitz, J.~L. Wright, and
  M.~Barnes, ``Situation awareness-based agent transparency and human-autonomy
  teaming effectiveness,'' \emph{Theoretical issues in ergonomics science},
  vol.~19, no.~3, pp. 259--282, 2018.

\bibitem{baraka2016expressive}
K.~Baraka, A.~Paiva, and M.~Veloso, ``Expressive lights for revealing mobile
  service robot state,'' in \emph{Robot 2015: Second Iberian Robotics
  Conference}.\hskip 1em plus 0.5em minus 0.4em\relax Springer, 2016, pp.
  107--119.

\bibitem{feng2018counterexamples}
L.~Feng, M.~Ghasemi, K.-W. Chang, and U.~Topcu, ``Counterexamples for robotic
  planning explained in structured language,'' in \emph{2018 IEEE International
  Conference on Robotics and Automation (ICRA)}.\hskip 1em plus 0.5em minus
  0.4em\relax IEEE, 2018, pp. 7292--7297.

\bibitem{kaptein2017personalised}
F.~Kaptein, J.~Broekens, K.~Hindriks, and M.~Neerincx, ``Personalised
  self-explanation by robots: The role of goals versus beliefs in robot-action
  explanation for children and adults,'' in \emph{2017 26th IEEE International
  Symposium on Robot and Human Interactive Communication (RO-MAN)}.\hskip 1em
  plus 0.5em minus 0.4em\relax IEEE, 2017, pp. 676--682.

\bibitem{gong2018behavior}
Z.~Gong and Y.~Zhang, ``Behavior explanation as intention signaling in
  human-robot teaming,'' in \emph{2018 27th IEEE International Symposium on
  Robot and Human Interactive Communication (RO-MAN)}.\hskip 1em plus 0.5em
  minus 0.4em\relax IEEE, 2018, pp. 1005--1011.

\bibitem{hastie2018miriam}
H.~Hastie, F.~J. Chiyah~Garcia, D.~A. Robb, A.~Laskov, and P.~Patron, ``Miriam:
  A multimodal interface for explaining the reasoning behind actions of remote
  autonomous systems,'' in \emph{Proceedings of the 20th ACM International
  Conference on Multimodal Interaction}, 2018, pp. 557--558.

\bibitem{quijano2017make}
L.~Quijano-Sanchez, C.~Sauer, J.~A. Recio-Garcia, and B.~Diaz-Agudo, ``Make it
  personal: a social explanation system applied to group recommendations,''
  \emph{Expert Systems with Applications}, vol.~76, pp. 36--48, 2017.

\bibitem{SocialMediaConflicts}
J.~M. {Such} and N.~{Criado}, ``Resolving multi-party privacy conflicts in
  social media,'' \emph{IEEE Transactions on Knowledge and Data Engineering},
  vol.~28, no.~7, pp. 1851--1863, 2016.

\bibitem{SocialMediaConflicts2}
R.~D. {Chhallani}, J.~{Rao}, and R.~{Pattewar}, ``Multi-party privacy conflict
  detection and resolution in social media,'' in \emph{2017 International
  Conference on Computing, Communication and Automation (ICCCA)}, 2017, pp.
  458--463.

\bibitem{SocialMediaConflicts3}
H.~Zhong, A.~Squicciarini, and D.~Miller, ``Toward automated multiparty privacy
  conflict detection,'' in \emph{Proceedings of the 27th ACM International
  Conference on Information and Knowledge Management}, ser. CIKM ’18, 2018,
  p. 1811–1814.

\bibitem{RecommendationSystemConflicts}
K.~McCarthy, J.~Reilly, L.~Mcginty, and B.~Smyth, ``Thinking positively -
  explanatory feedback for conversational recommender systems,'' 01 2004.

\bibitem{Pu2008UserInvolvedPE}
P.~Pu and L.~Chen, ``User-involved preference elicitation for product search
  and recommender systems,'' \emph{AI Magazine}, vol.~29, pp. 93--103, 2008.

\bibitem{sutton2018reinforcement}
R.~S. Sutton and A.~G. Barto, \emph{Reinforcement learning: An
  introduction}.\hskip 1em plus 0.5em minus 0.4em\relax The MIT Press, 2018.

\bibitem{daniel1990applied}
W.~W. Daniel \emph{et~al.}, ``Applied nonparametric statistics,'' 1990.

\bibitem{seltman2012experimental}
H.~J. Seltman, \emph{Experimental design and analysis}.\hskip 1em plus 0.5em
  minus 0.4em\relax Carnegie Mellon University Pittsburgh, 2012.

\bibitem{emirbayer_mische_1998}
M.~Emirbayer and A.~Mische, ``What is agency?'' \emph{American Journal of
  Sociology}, vol. 103, no.~4, p. 962–1023, 1998.

\end{thebibliography}

\end{document}